\pdfoutput=1

\documentclass[11pt]{article}

\usepackage[final]{acl}

\usepackage{times}
\usepackage{latexsym}
\usepackage{xcolor}
\usepackage{tabularx}
\usepackage{microtype}
\usepackage{enumitem}
\usepackage{float}

\usepackage[T1]{fontenc}

\usepackage[utf8]{inputenc}
\usepackage{amsmath}
\usepackage{graphicx}
\usepackage{amssymb}
\usepackage{dsfont}
\usepackage{algorithm}
\usepackage{algpseudocode}
\usepackage{booktabs}
\usepackage{array}
\usepackage{colortbl}
\usepackage{multirow}
\usepackage{multicol}
\definecolor{color3}{HTML}{FF8787}
\definecolor{color2}{HTML}{C7E4ED}
\definecolor{color1}{HTML}{E6EDCE}

\usepackage{microtype}

\usepackage{inconsolata}

\usepackage{graphicx}

\title{Improve Rule Retrieval and Reasoning with Self-Induction and Relevance ReEstimate}

\author{Ziyang Huang, Wangtao Sun, Jun Zhao, Kang Liu \\
        Institute of Automation, Chinese Academy of Sciences  \\
        University of Chinese Academy of Sciences \\
    \texttt{huangziyang2023@ia.ac.cn}
}

\begin{document}
\maketitle
\begin{abstract}
This paper systematically addresses the challenges of rule retrieval, a crucial yet underexplored area. Vanilla retrieval methods using sparse or dense retrievers to directly search for relevant rules to support downstream reasoning, often suffer from low accuracy. This is primarily due to a significant semantic gap between the instantiated facts in the queries and the abstract representations of the rules. Such misalignment results in suboptimal retrieval quality, which in turn negatively impacts reasoning performance. To overcome these challenges, we propose \textbf{Self-Induction Augmented Retrieval (SIAR)}, a novel approach that utilizes Large Language Models (LLMs) to induce potential inferential rules that might offer benefits for reasoning by abstracting the underlying knowledge and logical structure in queries. These induced rules are then used for query augmentation to improve retrieval effectiveness. Additionally, we introduce \textbf{Rule Relevance ReEstimate (R$^3$)}, a method that re-estimates the relevance of retrieved rules by assessing whether the abstract knowledge they contain can be instantiated to align with the facts in the queries and the helpfulness for reasoning. Extensive experiments across various settings demonstrate the effectiveness and versatility of our proposed methods.
\end{abstract}

\section{Introduction}

\begin{figure}[t]
    \centering
    \includegraphics[width=\linewidth]{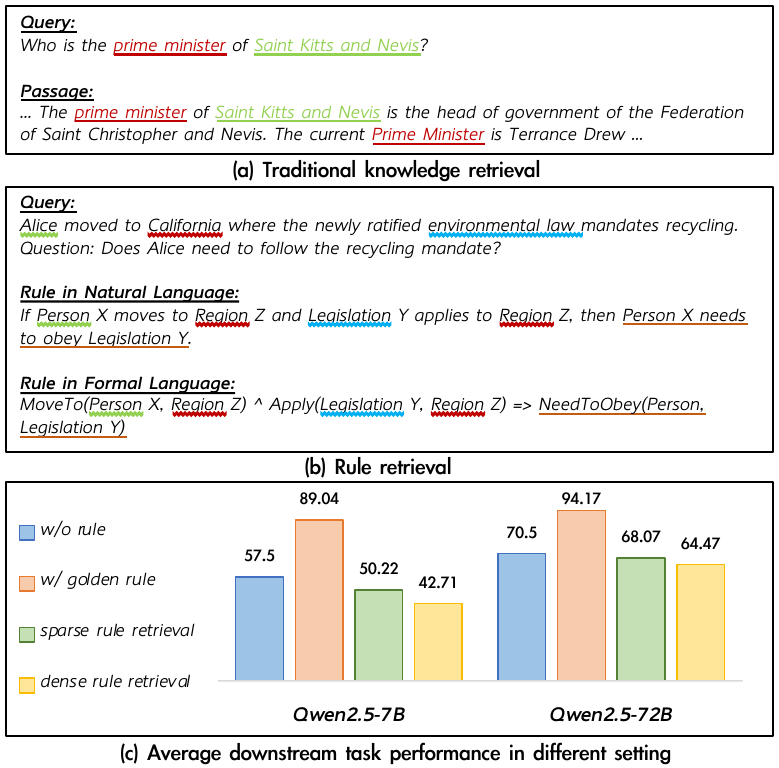}
    \caption{(a) and (b) show the different characteristics of traditional knowledge retrieval and rule retrieval. (c) illustrates that the golden rule can significantly improve reasoning performance, while existing rule retrieval methods typically lead to a decline in reasoning performance due to suboptimal recall.}
    \label{fig:intro}
\end{figure}
With the advancement of pre-training \cite{zhou2023comprehensivesurveypretrainedfoundation} and prompting techniques \cite{schulhoff2024promptreportsystematicsurvey, dong2024surveyincontextlearning}, Large Language Models (LLMs) \cite{zhao2024surveylargelanguagemodels, dubey2024llama3herdmodels, yang2024qwen2technicalreport, abdin2024phi3technicalreporthighly} have made significant progress in their understanding, reasoning, and decision-making capabilities \cite{Wang_2024}. Rules can generate new knowledge from existing information (or make decisions based on observed situations), which can enhance these abilities further \cite{zhu2024large, wang2024hypothesis}. 

The de facto approach of rule-based reasoning typically involves summarizing generalized rules from past experiences by LLMs, then retrieving the relevant rules based on the descriptions of downstream tasks or feedback from the observed environment, and finally using the retrieved rules to assist in reasoning or decision-making \cite{yang-etal-2023-failures, sun2023expnote, zhang-etal-2024-self-contrast, Zhao_Huang_Xu_Lin_Liu_Huang_2024}. Unfortunately, existing research has primarily focused on rule generation \cite{sivasothy2024largelanguagemodelsgenerating, wang-etal-2024-llms} and application \cite{wang2024symbolicworkingmemoryenhances}, neglecting the development of the rule retrieval techniques. Furthermore, rule retrieval plays a crucial role in real-world scenarios. For example, in legal scenarios, one must retrieve the relevant laws based on the crime to make a judgment \cite{xiao2018cail2018largescalelegaldataset}, and in medical settings, domain-specific rules must be retrieved based on symptoms to assist in diagnosis \cite{wang2024rulealign}. The above highlights the urgency of exploring rule retrieval area.

The rules discussed in this paper are referential rules \cite{sun2024beyond, wang-etal-2024-llms}, which typically manifest as the derivation of one set of facts from another. In natural language, they are usually expressed in the form "if \texttt{Premise}, then \texttt{Conclusion}," whereas in formal language, they are often represented as "\texttt{Premise} $\Rightarrow$ \texttt{Conclusion}." As illustrated in Figure \ref{fig:intro} (a), the query and the corresponding golden passage typically share some keywords or similar semantics in traditional knowledge retrieval scenarios \cite{bajaj2016ms, karpukhin-etal-2020-dense}, making direct matching between the query and passage feasible. However, in rule retrieval scenarios, the following characteristics of the query and golden rule present significant challenges: (1) The facts in the query are instantiated and specific.
(2) The facts contained in the rule are composed of variables and predicates, where the variables typically have an abstract, conceptual type, and the predicates represent relationships between different variables, which might not be expressed in the query explicitly.
(3) The gap between the query (concrete, implicit) and the rule (abstract, explicit) leads to a semantic misalignment. (4) There is an explicit derivation in the rule, but not in the query. For example, as shown in Figure \ref{fig:intro} (b), "environmental law in California mandates recycling" is the fact in the query. In the golden rule, the entity "California" corresponds to "Region Z", "environmental law" corresponds to "Legislation Y", and this fact implies "Legislation Y applies to Region Z". Furthermore, the rule incorporates the inferred conclusion "Person X needs to obey Legislation Y".

Existing methods typically overlook the aforementioned characteristics and apply traditional retrieval techniques directly during the rule retrieval phase. As demonstrated in Figure \ref{fig:intro} (c), whether using sparse retrieval or dense retrieval, relying on vanilla retrieval to assist in reasoning often results in varying degrees of performance degradation compared to reasoning without rules. This demonstrates the inadequacy of traditional retrieval methods in rule-based scenarios. In fact, if the retrieved rules are irrelevant or contain noise, the reasoning will be distracted. As the size of the rule base increases, rule retrieval will become the bottleneck for downstream task performance. In contrast, when the golden rule is directly provided to aid the LLM in its reasoning process, performance enhancement of 31.54\% and 23.67\% can be witnessed in the 7B and 72B models respectively. The performance gap highlights the importance of rules in supporting reasoning and the necessity of accurate rule retrieval.


To this end, this paper proposes \textbf{Self-Induction Augmented Retrieval (SIAR)}. SIAR leverages self-induction to summarize and abstract the facts presented in the query and hypothesize potential inferential relationships to generate a potential rule. This newly generated rule is then used as the new query, or combined with the original query to form a new query for retrieval. Specifically, we utilize few-shot in-context learning to prompt the LLM to produce a self-induced rule. Our theoretical insight is as follows: if we consider the query set and the rule set as belonging to different semantic subspaces, where the former is characterized by instantiated, concrete facts and the latter by abstract, conceptual knowledge. We hypothesize that these two subspaces are nearly non-overlapping. The role of self-induction is to project the query as much as possible into the rule subspace, enabling the query to better match rules that share similar underlying logic during retrieval.

Although SIAR can improve the ranking of the golden label in the retrieved rule list, the limited inductive capabilities of LLMs still make it challenging to handle more difficult queries. Moreover, the retriever can only evaluate the semantic similarity instead of the helpfulness of the rule for the query. Therefore, building on SIAR, we propose \textbf{Rule Relevance ReEstimate (R$^3$)}, which utilizes the LLMs to estimate the relevance of the retrieved rule list. R$^3$ evaluates whether each rule can be applied to the current query for better reasoning and reranks the list based on the relevance estimation.

 We conduct experiments on two synthetic datasets as well as one real-world dataset. Compared to direct retrieval, SIAR achieves significant improvements in both retrieval and reasoning performance, demonstrating its effectiveness in extracting and summarizing the knowledge and logic embedded in queries to assist retrieval. Building on SIAR, R$^3$ further enhances both retrieval and reasoning performance, proving that LLMs can reliably assess the relevance between queries and rules, thereby improving the quality of rule retrieval. Moreover, SIAR and R$^3$ consistently improve performance across different settings, including varying rule formats (natural and formal language), different types of retrievers (sparse and dense retrievers), and LLMs of different parameter scales. This demonstrates the generalizability of our proposed methods.

The contributions of this paper are as follows:
\begin{itemize}
    \item We systematically introduce the problem of rule retrieval and provide a detailed analysis of the semantic misalignment challenges faced in rule retrieval. 
    \item We propose SIAR and R$^3$ to address the issues in rule retrieval: SIAR induces and abstracts the knowledge and logic embedded in the query to map it into the rule space for more effective retrieval, while R$^3$ enhances retrieval quality by assessing the relevance of retrieved rules to the original query.
    \item Extensive experiments demonstrate that SIAR and R$^3$ achieve performance improvements across various datasets and settings. Furthermore analysis offers more insights for future research on rule retrieval.

\end{itemize}

\begin{figure*}[t]
    \centering
    \includegraphics[width=\linewidth]{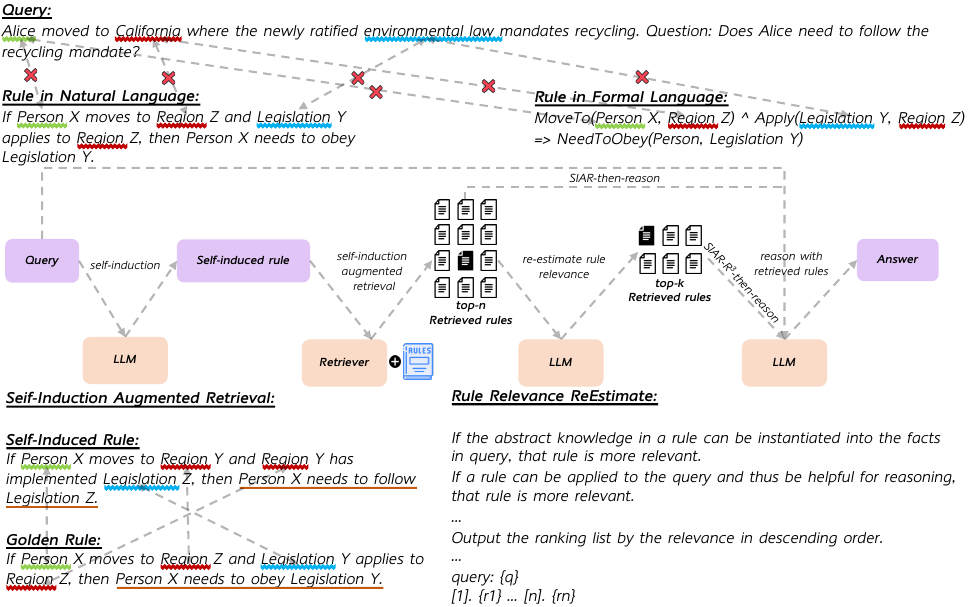}
    \caption{The workflow of \texttt{retrieve-then-reason} augmented with our method is shown in the middle of the Figure. To address the semantic misalignment issues, self-induction is first utilized to generate the hypothesized rule for query augmentation. Then, the new query is used for rule retrieval. And the retrieved rules are concatenated with the original query for reasoning. Building on this, we can reestimate the relevance of the rules with the query and improve the retrieval quality for better reasoning. The left bottom of the Figure shows the example of the self-induced rule. And the right bottom of the Figure shows the simplified reestimation prompt.}
    \label{fig:method}
\end{figure*}

\section{Preliminary}
In the rule reasoning scenario, we have a rule library $\mathcal{R} = \{r_i\}_{i=1}^{|\mathcal{R}|}$, which incorporates inferential rules offering benefits for new knowledge induction and decision-making. We use the query $q$ to retrieve relevant rules from this library, and the retrieved rules are concatenated with the query as context. This combined input is then fed into the LLM to perform reasoning. We name this workflow as \texttt{retrieve-then-reason}.

In the retrieval phase, we employ either sparse retrieval or dense retrieval. The former uses BM25 to compute the similarity between the query and the rules, while the latter leverages a pre-trained encoder to map both the query and the rules into a shared vector space, where their similarity is computed using the cosine function. The top-k rules are then returned based on the similarity ranking. To improve retrieval speed, we pre-build and cache the index of $\mathcal{R}$.

\section{Method}
In the aforementioned \texttt{retrieve-then-reason} paradigm, our method primarily focuses on improving retrieval quality by aligning the semantics between the query and the rule and re-assessing the relevance of retrieved rules. The former technique is inserted before the retrieval stage and the latter one is inserted between retrieval and reasoning. Furthermore, we aim to positively impact the overall reasoning performance. Our method is based on prompting the LLMs without any training, which is versatile and cost-efficient.

\subsection{SIAR: Self-Induction Augmented Retrieval}
As illustrated in Figure \ref{fig:method}, before performing rule retrieval, we employ a self-induction process where the LLM generates a potentially useful rule to aid reasoning. This process highly relies on the inductive capability of LLM \cite{wang2024hypothesis, zhu2024large, bowen-etal-2024-comprehensive, cheng2024inductive}. We refer to this generated rule as the self-induced rule (SI). The key to self-induction lies in summarizing and abstracting the facts embedded in the query and hypothesizing potential inferential relationships. Due to the different characteristics of query and rule, the primary target of self-induction is to project the query to rule space. Specifically, we utilize few-shot prompting to guide the LLM for self-induction, with the corresponding instruction template shown in Appendix \ref{app:prompt}. We show one self-induction example in Figure \ref{fig:method}.

After generating the SI, we have two options for utilizing it in the retrieval process: we can either treat the SI as the new search query, or we can concatenate the SI with the original query to form a new combined search query. The former approach is referred to as SIAR (w/ SI), while the latter is referred to as SIAR (w/ SI + input). These two designs result from the different natures of sparse retrieval and dense retrieval, and we talk about the impact in Section \ref{sec:experiment}.

\subsection{R$^3$: Rule Relevance ReEstimate}

The principle of retrieval is to match two different strings based on keyword or semantic similarity. Therefore, if the inductive capability of the LLM is not strong enough, the generated rule might still not align well with the golden rule. As a result, the retrieval list produced by SIAR may still have suboptimal ranking quality. Moreover, the retriever cannot determine whether a rule can aid the LLM in reasoning, nor can it specifically assess the relationship between the two. To address these, we propose to rerank the top-n rules from the previous stage by evaluating the relevance of the retrieved rules to the original query, as shown in Figure \ref{fig:method}. The key to R$^3$ is determining whether the abstract knowledge in a rule can be instantiated into the facts contained in the query, and whether the rule can assist the LLM in reasoning.

Inspired by RankGPT \cite{sun-etal-2023-chatgpt}, we prompt the LLM to directly output a ranked list of rules, which can reduce the prompting times compared to pair-wise estimation and thus accelerate the R$^3$ process. And we select the top-k rules from this reranked list as the final retrieval result. The corresponding instruction template is shown in Appendix \ref{app:prompt}. This prompt encourages the LLM to assess both the relevance and utility of each rule, ensuring a more accurate final retrieval. Based on the query used in SIAR, R$^3$ also has two versions: R$^3$ (w/ SI) and R$^3$ (w/ SI + input).

\begin{table*}[t]
\centering
\tabcolsep=15pt
\resizebox{1\linewidth}{!}{
\begin{tabular}{lccccccccc}
\toprule
\multirow{2}{*}{} & \multicolumn{3}{c}{\textbf{CLUTRR}} & \multicolumn{3}{c}{\textbf{ULogic}}  & \multicolumn{3}{c}{\textbf{CAIL2018}}  \\
\cmidrule(l){2-10}
 & R@1 & R@5  & R@10 & R@1 & R@5  & R@10 & R@1 & R@5  & R@10  \\
\cmidrule{1-10}
\textbf{sparse retrieval (BM25)} \\
\cmidrule(l){1-1}
\texttt{vanilla retrieval} &  6.67 & 16.60 & 24.52 & 68.91 & 85.42 & 92.29 & 25.30 & 49.40 & 59.04  \\
\cmidrule{1-10}
\texttt{gpt-4o}  \\
+ SIAR (w/ SI) &  7.16 & 15.55 & 22.71 & 58.43 & 81.08 & 87.35 & 68.07 & 87.34 & 93.98  \\ 
+ SIAR (w/ SI + input) & 10.78 & 23.00 & 30.73 & 75.78 & 91.93 & 96.63 & 61.45 & 84.33 & 89.16 \\ 
+ SIAR-R$^3$ (w/ SI) &  11.45 & 19.75 & 24.62 & 87.83 & 92.65 & 92.65 & 83.13 & \textbf{93.38} & \textbf{93.98} \\ 
+ SIAR-R$^3$ (w/ SI + input) &  \textbf{16.32} & \textbf{26.81} & \textbf{32.92} & \textbf{92.65} & \textbf{97.71} & \textbf{97.83} & \textbf{83.73} & 92.17 & 92.17 \\ 
\cmidrule{1-10}
\texttt{Qwen2.5-72B-Instruct}  \\
+ SIAR (w/ SI) &  8.11 & 17.36 & 26.04 & 57.22 & 79.76 & 86.87 & 78.91 & 88.55 & 90.96  \\ 
+ SIAR (w/ SI + input) & 11.06 & 23.66 & \textbf{32.44} & 74.82 & 90.48 & 94.94 & 74.70 & 86.14 & 89.76 \\ 
+ SIAR-R$^3$ (w/ SI) &  13.36 & 22.42 & 28.24 & 87.47 & 93.25 & 93.37 & \textbf{86.75} & \textbf{93.98} & \textbf{93.98} \\ 
+ SIAR-R$^3$ (w/ SI + input) &  \textbf{14.31} & \textbf{25.38} & 31.58 & \textbf{92.17} & \textbf{96.39} & \textbf{96.75} & 86.14 & 92.17 & 92.17 \\ 
\cmidrule{1-10}
\texttt{Qwen2.5-7B-Instruct}  \\
+ SIAR (w/ SI) &  2.29 & 8.30 & 11.93 & 60.48 & 83.86 & 90.00 & 62.05 & 71.08 & 77.11  \\ 
+ SIAR (w/ SI + input) & \textbf{7.06} & \textbf{16.70} & \textbf{22.81} & 76.14 & 90.24 & 95.18 & 57.83 & 74.10 & 78.31 \\ 
+ SIAR-R$^3$ (w/ SI) &  2.00 & 6.39 & 10.02 & 84.81 & 93.13 & 93.37 & 72.89 & 80.12 & 80.12 \\ 
+ SIAR-R$^3$ (w/ SI + input) &  4.58 & 12.5 & 19.27 & \textbf{88.67} & \textbf{96.14} & \textbf{96.50} & \textbf{75.30} & \textbf{83.13} & \textbf{84.33} \\ 
\cmidrule{1-10}
\textbf{dense retrieval (bge)} \\
\cmidrule(l){1-1}
\texttt{vanilla retrieval} &  2.10 & 7.73 & 12.02 & 30.36 & 58.43 & 71.45 & 9.04 & 15.66 & 21.69  \\
\cmidrule{1-10}
\texttt{gpt-4o}  \\
+ SIAR (w/ SI) &  12.31 & 22.61 & 28.91 & 65.18 & 87.11 & 91.33 & 56.02 & 74.70 & 81.33  \\ 
+ SIAR (w/ SI + input) & 5.25 & 12.60 & 19.37 & 43.61 & 73.13 & 83.61 & 21.68 & 40.96 & 51.81 \\ 
+ SIAR-R$^3$ (w/ SI) & \textbf{16.51} & \textbf{25.38} & 30.34 & \textbf{86.62} & \textbf{94.58} & \textbf{94.70} & \textbf{78.31} & \textbf{84.94} & \textbf{84.94} \\ 
+ SIAR-R$^3$ (w/ SI + input) &  10.21 & 18.80 & \textbf{31.58} & 80.72 & 88.92 & 89.52 & 55.42 & 60.84 & 60.84 \\ 
\cmidrule{1-10}
\texttt{Qwen2.5-72B-Instruct}  \\
+ SIAR (w/ SI) &  11.74 & \textbf{24.71} & \textbf{31.58} & 64.82 & 86.74 & 92.65 & 76.51 & 84.34 & 85.54  \\ 
+ SIAR (w/ SI + input) & 4.68 & 12.98 & 19.37 & 41.80 & 71.80 & 83.25 & 23.49 & 49.40 & 57.23 \\ 
+ SIAR-R$^3$ (w/ SI) & \textbf{14.03} & 23.09 & 30.25 & \textbf{88.19} & \textbf{94.70} & \textbf{95.06} & \textbf{81.32} & \textbf{88.55} & \textbf{89.15} \\ 
+ SIAR-R$^3$ (w/ SI + input) & 10.31 & 16.22 & 20.32 & 83.86 & 89.76 & 90.00 & 66.27 & 69.28 & 69.28 \\ 
\cmidrule{1-10}
\texttt{Qwen2.5-7B-Instruct}  \\
+ SIAR (w/ SI) &  \textbf{5.53} & \textbf{12.21} & \textbf{16.13} & 71.57 & 90.96 & 95.54 & 59.64 & 68.07 & 69.88  \\ 
+ SIAR (w/ SI + input) & 2.96 & 10.88 & 16.13 & 42.89 & 71.93 & 83.01 & 23.49 & 43.98 & 51.81 \\ 
+ SIAR-R$^3$ (w/ SI) & 2.39 & 7.73 & 11.45 & \textbf{87.59} & \textbf{96.39} & \textbf{97.11} & \textbf{70.48} & \textbf{74.10} & \textbf{74.70} \\ 
+ SIAR-R$^3$ (w/ SI + input) & 2.29 & 7.54 & 11.64 & 76.75 & 87.95 & 88.92 & 56.02 & 59.64 & 60.24 \\ 
\bottomrule
\end{tabular}
}
\caption{Performance of different methods with rule library in \textit{Natural Language}. We use Recall@1, Recall@5 and Recall@10 as the retrieval metrics.}
\label{tab:natural_retrieval}
\end{table*}

\section{Experiment}
\label{sec:experiment}
We select two synthetic datasets, Clutrr \cite{sinha-etal-2019-clutrr} and ULogic \cite{wang-etal-2024-llms}, as well as a real-world dataset, CAIL2018 \cite{xiao2018cail2018largescalelegaldataset} from RuleBench for our evaluation. We report Recall@1, Recall@5, Recall@10 for retrieval results and Match for reasoning results. We use gpt-4o \cite{gpt4o} and Qwen2.5 \cite{qwen2.5} series (7B and 72B) as the tested LLMs. Due to space limitation, we put the entire experiment setting in Appendix \ref{experiment setting}.

\begin{table*}[tb]
    \centering
    \resizebox{1\linewidth}{!}{
    \begin{tabular}{lccccccccc}
    \toprule
    & \multicolumn{3}{c}{\texttt{gpt-4o}} & \multicolumn{3}{c}{\texttt{Qwen2.5-72B-Instruct}} & \multicolumn{3}{c}{\texttt{Qwen2.5-7B-Instruct}} \\
    \cmidrule{2-10}
    & CLUTRR & ULogic & CAIL2018 & CLUTRR & ULogic & CAIL2018 & CLUTRR & ULogic & CAIL2018 \\
    \midrule
    \textit{w/o retrieval} \\
    Direct & 42.65 & 92.28 & 76.47 & 38.36 & 93.01 & 80.12 & 25.34 & 87.47 & 61.45 \\
    Golden rule & 93.51 & 89.40 & 98.67 & 89.03 & 94.58 & 98.90 & 82.06 & 88.67 & 96.39 \\
    CoT & 51.34 & 93.61 & 77.85 & 49.43 & 90.12 & 83.13 & 17.84 & 88.07 & 69.88\\
    Self-Induction & 50.76 & 87.83 & 82.98 & 49.62  & 91.69 & 84.94 & 31.58  &88.43 & 64.46 \\
    \midrule
    \textit{w/ sparse retrieval} \\
    vanilla & 37.69 & 89.04 & 74.36 & 37.60 & 93.13 & 73.49 & 26.81 & 87.11 & 36.75 \\
    SIAR & 46.09 & 87.71 & 80.77 & 49.14 & 94.21 & 86.14 & 33.87 & 88.92 & 59.64 \\
    SIAR-R$^3$  & 49.33  & \textbf{89.64} & \textbf{85.71} & \textbf{51.71}  & \textbf{95.90} & \textbf{86.75} & \textbf{33.97}  & 91.33 & \textbf{73.49} \\
    \midrule
    \textit{w/ dense retrieval} \\
    vanilla & 34.06 & 88.07 & 80.65 & 30.53 & 90.00 & 72.89 & 25.00 & 83.25 & 19.88 \\
    SIAR & 52.19 & 86.39 & 80.75 & 49.81 & 95.06 & 86.75 & 34.73 & 89.64 & 60.24 \\
    SIAR-R$^3$  & \textbf{54.20}  & 89.28 & 83.54 & 51.05  & 95.78 & 84.94 & 33.59  & \textbf{91.81} & 68.07 \\
    \bottomrule
    \end{tabular}
    }
    \caption{Downstream reasoning performance. We use Match as the metric.}
    \label{tab:generation}
\end{table*}

\subsection{Retrieval Results and Discussion}
\label{sec:4.1}
As shown in Table \ref{tab:natural_retrieval} and Table \ref{tab:formal_retrieval}, we report the retrieval performance using the different rule libraries (natural vs. formal). In each table, we present the performance of different retrievers (sparse vs. dense), LLMs with varying architectures (openai-gpt vs. Qwen) and parameter scales (72B vs. 7B), and different forms of queries (w/ SI vs. w/ SI + input). Due to space limitation, we put the formal language results in the Table \ref{tab:formal_retrieval} of the Appendix \ref{formal retrieval results}.

\paragraph{Open-source models have comparable performance with closed-source models.}
In most settings, Qwen2.5-72B-Instruct demonstrates performance similar to GPT-4o, and in certain configurations, such as SIAR-R3 (w/ SI) + CAIL2018, it even outperforms GPT-4o. Moreover, compared to the baseline, they all achieve better performance. Therefore, we believe that open-source models have reached a level of rule induction and ranking capability comparable to that of the most advanced closed-source models.
For fair comparison, the following analysis uses 72B and 7B Qwen models within the same family.

\paragraph{SIAR can consistently improve performance compared to vanilla retrieval.}
Under various combinations of retrievers, rule formats, and query formats, SIAR consistently outperforms direct retrieval. In different scenarios, SIAR achieves improvements of up to 9.64 (natural, dense, 72B, w/ SI), 60.12 (formal, dense, 72B, w/ SI), and 67.47 (natural, dense, 72B, w/ SI) in Recall@1 on Clutrr, ULogic, and CAIL2018, respectively. These results highlight the self-induction capabilities of 
LLMs, enabling them to effectively project queries into the rule space and reduce semantic misalignment between queries and rules. Additionally, we observe that models with 72B parameters tend to exhibit greater performance gains compared to 7B models, suggesting that inductive abilities improve with larger model scales.

\paragraph{SIAR-R$^3$  can usually improve performance compared to SIAR.}
On ULogic and CAIL2018, R$^3$ significantly boosts the performance of SIAR across all setup combinations. Notably, SIAR-R$^3$ achieves maximum improvements in Recall@1 of 43.25 (formal, dense, 72B, w/SI + input) and 42.78 (natural, dense, 72B, w/SI + input). These results indicate that R$^3$ effectively reevaluates and reranks the relevance of rules retrieved by SIAR. By directly assessing the relevance between the query and the rule, R$^3$ overcomes the limitations of retrievers that rely solely on keyword or semantic similarity, thus enhancing retrieval quality. Additionally, on the CLUTRR dataset, performance gains were only observed in models with 72B parameters, and the improvements from the formal rule base were smaller than those from the natural language rule base. This suggests that on more complex datasets, models with smaller parameter scales lack the capacity to effectively rerank rules, limiting their ability to drive performance improvements.

\paragraph{The performance difference of sparse retrieval and dense retrieval depends on the format of rule and dataset.}
On the Clutrr and CAIL2018 datasets, sparse retrieval generally outperforms dense retrieval. However, on the ULogic dataset, performance varies depending on the rule base used. With the natural language rule base, sparse retrieval achieves a higher accuracy (92.17) compared to dense retrieval (88.19). Conversely, with the formal rule base, dense retrieval (89.75) surpasses sparse retrieval (80.60). This suggests that retrieval performance is highly dependent on the dataset and the linguistic form of the rule base. Despite these variations, we believe that in most cases, sparse retrieval will outperform dense retrieval. This is because, in rule-based scenarios, many concepts may not be well-represented in dense vector spaces. In contrast, sparse retrieval, which relies on keyword matching, may offer a more precise alignment between the query and the corresponding rules.

We add more analysis in Appendix \ref{retrieval analysis}.

\begin{table*}[htb]
    \centering
    \resizebox{1\linewidth}{!}{
    \begin{tabular}{lccccccccc}
    \toprule
    & \multicolumn{3}{c}{\texttt{Qwen-2.5-7B-Instruct}} & \multicolumn{3}{c}{\texttt{Llama-3.1-8B-Instruct}} & \multicolumn{3}{c}{\texttt{Yi-1.5-6B-Chat}} \\
    \cmidrule{2-10}
    & R@1 & R@5 & R@10 & R@1 & R@5 & R@10 & R@1 & R@5 & R@10 \\
    \midrule
    \textit{w/ sparse retrieval} \\
    vanilla retrieval & 68.91 & 85.42 & 92.29 & 68.91 & 85.42 & 92.29 & 68.91 & 85.42 & 92.29 \\
    vanilla retrieval + R$^{3}$ & 86.99& 93.98& 94.70& 56.71& 92.77& 93.98 & 69.88 & 86.14 & 91.08 \\
    SIAR (w/ SI) & 71.57 & 90.96& 95.54&54.94& 80.12& 86.87&59.76& 82.65& 88.31 \\
    SIAR (w/ SI + input) & 42.89& 71.93& 83.01&\textbf{76.87}& \textbf{91.69}& \textbf{96.27}& \textbf{74.58}& \textbf{90.10}& \textbf{95.06} \\
    SIAR-R$^3$ (w/ SI) & 84.81& 93.13& 93.37&61.08& 89.04& 90.36&61.08& 83.73& 87.83\\
    SIAR-R$^3$ (w/ SI + input) & \textbf{88.67}& \textbf{96.14}& \textbf{96.50}&58.31& 94.94& 96.62&73.25& 90.00& 94.22 \\
    \midrule
    \textit{w/ dense retrieval} \\
    vanilla retrieval & 30.36& 58.43& 71.45&30.36& 58.43& 71.45&30.36& 58.43& 71.45 \\
    vanilla retrieval + R$^{3}$ & 70.12& 79.88& 80.48&59.15& 78.43& 79.64&39.52& 61.93& 71.33 \\
    SIAR (w/ SI) & 60.48& 83.86& 90.00&59.76& 85.54& 90.36&\textbf{69.28}& \textbf{87.95}& \textbf{92.05}\\
    SIAR (w/ SI + input) & 76.14& 90.24& 95.18&42.29& 72.29& 82.53&41.45& 70.60& 82.05 \\
    SIAR-R$^3$ (w/ SI) & \textbf{87.59}& \textbf{96.39} & \textbf{97.11} & \textbf{66.75}& \textbf{92.17}& \textbf{93.01}&68.19& 87.59& 91.32\\
    SIAR-R$^3$ (w/ SI + input) &76.75& 87.95& 88.92&63.86& 87.35& 89.40&47.71& 71.80& 81.69 \\
    \bottomrule
    \end{tabular}
    }
    \caption{Retrieval performance with different types of models.}
    \label{tab:models}
\end{table*}

\subsection{Reasoning Results and Discussion}
\paragraph{Baselines} (1) Direct: answer the question directly. This is set as the bottom of the performance. (2) Golden rule: answer the question with the golden rule. This is set as the ceil of performance. (3) CoT \cite{wei2022chain}: reason step by step and then produce the answer. (4) Self-Induction: answer the question with the self-induced rule. (5) vanilla retrieval: use the original query to retrieve the rule and then answer the question.

Based on the conclusion from the previous section, for sparse retrieval, we use SI+input as the query, while for dense retrieval, we use SI as the query for retrieval. And we use the rule library in natural language. We report downstream reasoning performance in Table \ref{tab:generation}.

We use the average performance enhancement over three different datasets to analyze and get the following conclusions.
Similarly as the Section \ref{sec:4.1}, GPT-4o and Qwen2.5-72B have comparable performance, so we use Qwen-72B and Qwen-7B for further analysis.

An exception occurs with the ULogic, where gpt-4o outperforms the golden rule even without utilizing the rules. Based on our observations, gpt-4o has already achieved a relatively saturated performance (>90\%) on this dataset, and additional rule knowledge may not bring further performance improvements on this dataset. Apart from this, the conclusions drawn from the analysis are reflected on the other two datasets.

\paragraph{Rules can effectively assist LLMs in reasoning, while directly retrieving rules for reasoning may lead to a decline in performance.}

Incorporating the Golden Rule as an aid in reasoning, rather than directly answering questions, has significantly improved performance across various models. For instance, the Qwen2.5-7B-Instruct model saw an average improvement of 31.54, while the Qwen2.5-72B-Instruct model showed a gain of 23.67. These substantial improvements suggest that the Golden Rule effectively enhances the ability of LLMs to infer from existing information, generate new knowledge, and make more reasonable decisions. In contrast, when relying on vanilla retrieval, performance decreases by 7.28 on the Qwen2.5-7B-Instruct model and by 2.43 on the Qwen2.5-72B-Instruct model. Vanilla dense retrieval leads to even larger drops, with declines of 14.79 and 6.03, respectively. These findings indicate that reasoning without accurate rule-based assistance, such as the golden rule, is less effective when based solely on vanilla retrieved results. It is noteworthy that using the question directly as a query for rule retrieval often produces low-quality, noisy results. The noise negatively impacts the reasoning process and degrades model performance. This phenomenon underscores a key challenge faced by current retrieval systems: semantic misalignment between queries and rules. Existing retrieval techniques struggle to accurately compute the similarity between the two, resulting in difficulty retrieving truly relevant rules, which ultimately hampers reasoning performance.

\paragraph{SIAR and R$^3$ can boost the performance significantly.}

The SIAR method significantly enhances model performance compared to direct retrieval. 
In scenarios utilizing sparse retrieval, performance increased by 31.76 and 25.27 for the 7B and 72B models, respectively. The improvements are even more pronounced with dense retrieval, where performance gains reached 56.48 and 38.20 for the 7B and 72B models. These results demonstrate that SIAR provides substantial performance boosts across models of varying sizes. When the R$^3$ mechanism was introduced, performance improved further. In sparse retrieval, the 7B model gains an additional 16.36, while the 72B model sees an increase of 4.87 points. For dense retrieval, the 7B model achieves an extra gain of 8.86, and the 72B model improves by 0.15. These findings validate the effectiveness of SIAR in enhancing retrieval quality, allowing for better alignment between queries and relevant rules, which in turn strengthens the reasoning process. 
SIAR addresses the semantic mismatch inherent in traditional retrieval methods by self-induction to map queries into the rule space.
The R$^3$ mechanism further refines the retrieval by reassessing the relevance and applicability of each rule to the current query, overcoming the limitations of traditional retrievers that struggle to evaluate rules effectively.
Compared to other baselines that do not utilize retrieval, our method demonstrates significant superiority. These results highlight the critical role of high-quality rule retrieval in reasoning tasks, showing that accurate retrieval is essential for improving reasoning performance.

\section{Ablation Study}
We perform more ablation experiments to explore more influencing factors.
We use the ULogic dataset and test six methods: vanilla prompt, vanilla prompt + R$^3$, SIAR (w/ SI), SIAR (w/ SI + input), SIAR-R$^3$ (w/ SI), and SIAR-R$^3$ (w/ SI + input). Among them, "vanilla prompt + R$^3$" refers to retrieving using the original query and then directly performing R$^3$. Due to space limitations, this method was not presented in the previous section. Moreover, we put the result table of Section \ref{abl: retriever} and Section \ref{abl: number of rule} in Appendix \ref{abl: app}.

\subsection{The effects of different models}
Different LLMs have different model architectures and use different training data. To demonstrate the generalizability of our method, we conducted experiments on a wider range of model types \cite{dubey2024llama3herdmodels, ai2025yiopenfoundationmodels}, as shown in the table \ref{tab:models}. The results show that our method achieves a significant improvement over the baseline across different models.

\subsection{The effects of different retrievers}
\label{abl: retriever}
Different types of retrievers have different characteristics. To validate the generalizability of our method, we compared the performance of three different types of retrievers: sparse retriever (bm25), dense retriever (bge), and LLM retriever (bge-gemma2 \cite{chen2024bgem3embeddingmultilingualmultifunctionality}). For comparison, the dense retriever has only 110M parameters, and the LLM retriever has 9B parameters. As shown in the table \ref{tab:retrievers}, the results demonstrate that our method performs well across different types of retrievers. Even with large retrieval model, our method is still able to provide further enhancement, which strongly demonstrates the generalizability of our approach.

\subsection{The effects of the number of rules}
\label{abl: number of rule}
 To validate the robustness of our method, we added the counterfactual rule set from the ulogic dataset (constructed by the original RuleBench \cite{sun2024beyond}) to the original rule set and re-tested the performance of our method, as shown in Table \ref{tab:models}. In this setup, the number of rules doubles compared to the original. As the number of irrelevant rules increases, the performance of retrieval will continuously decline. So the number of rules is a very important influencing factor. However, our method still demonstrates a significant improvement compared to the baseline.

\section{Related Work}

\subsection{LLM and rule}
As the inductive \cite{yang-etal-2024-language, wang-etal-2024-llms} and deductive \cite{saparov2023testinggeneraldeductivereasoning} capabilities of LLMs continue to advance, they are increasingly being employed to summarize latent transformation patterns from sets of inputs and outputs \cite{sun-etal-2024-itd, qiu2024phenomenal}. These patterns are then formalized as executable rules, stored, and used to support reasoning in downstream tasks \cite{yang-etal-2023-failures, sun2023expnote, zhu2024large, wang2024hypothesis, wang2024symbolicworkingmemoryenhances}.

More specifically, They learn rules from input-output pairs to represent relationships between inputs and outputs, then use these rules for reasoning and quality verification. High-quality rules are stored in a library \cite{zhu2024large, wang2024hypothesis}. Previous research on rule retrieval includes two methods: one concatenates all rules with the input for inference, requiring a hierarchical storage structure  \cite{zhu2024large}; the other uses vanilla retrieval \cite{sun2023expnote, yang-etal-2023-failures}, which deteriorates as the rule set grows. This paper focuses on addressing semantic misalignment and relevance estimation issues in retrieval, proposing solutions to improve semantic matching and relevance evaluation for more accurate rule retrieval.

\subsection{Generation Augmented Retrieval}

The Generation Augmented Retrieval (GAR) \cite{mao-etal-2021-generation} is a common approach that leverages the capabilities of language models to perform query decomposition \cite{chen-etal-2024-analyze}, query rewriting \cite{ma-etal-2023-query}, and query expansion \cite{wang-etal-2023-query2doc}, helping to supplement missing background knowledge in queries to achieve higher retrieval quality. In addition to passage retrieval, GAR can also play a role in code retrieval \cite{li-etal-2024-rewriting}.  Our SIAR can be seen as a type of GAR that utilizes the self-inductive abilities of large language models (LLMs).
\section{Conclusion}
This paper introduces Self-Induction Augmented Retrieval (SIAR) and Rule Relevance Re-Estimate (R$^3$) to address the challenges of rule retrieval in complex reasoning tasks. These techniques significantly enhance retrieval accuracy by LLMs to induce abstract inferential rules and assess the relevance of retrieved rules to queries.
SIAR and R3 offer promising solutions for overcoming the semantic misalignment issues in traditional retrieval techniques, paving the way for more effective rule-based reasoning in real-world applications.
\section*{Limitations}
Currently, the rule libraries we discussed remain quite limited in size, as seen in datasets like Clutrr, ULogic, and CAIL2018, which contain only 1,048, 830, and 166 rules, respectively. Compared to the vast number of articles in traditional passage retrieval, the rule bases we retrieved are still relatively small. However, even with these small datasets, traditional retrieval methods have shown a decline in reasoning performance, underscoring the need for deeper exploration in rule retrieval. The smaller number of rules reduces the difficulty of the benchmark. In future work, we aim to introduce more irrelevant rules to explore additional challenges in rule retrieval.
\bibliography{custom}

\appendix
\newpage
\section{Prompt Template}
\label{app:prompt}

\begin{table}[h]
\begin{tabularx}{\linewidth}{X}
\toprule 
\textbf{Self-Induction Prompt Template} \\
\cmidrule{1-1}
You are given a Query. Please write the inferential rule may help answer the question. The rule should summarize and abstract the facts in the query and catch the underlying logic. I will give some examples. Just output the rule and do not output anything else.\\
query:  \{\textit{q$_1$}\} \\
rule: \{\textit{r$_1$}\} \\
... more demonstrations
\\
\bottomrule

\end{tabularx}

\end{table}

\begin{table}[h]
\begin{tabularx}{\linewidth}{X}
\toprule 
\textbf{Rule Relevance ReEstimate Prompt Template} \\
\cmidrule{1-1}
You are an intelligent assistant that can rank rules based on their relevancy to the query. If the abstract knowledge in a rule can be instantiated into the facts in query, that rule is more relevant.
If a rule can be applied to the query and thus be helpful for reasoning, that rule is more relevant. \\
I will provide you with \{\textit{num\_rule}\} rules, each indicated by a numerical identifier []. Rank the rules based on their relevance to the query: \{\textit{query}\}. \\
 \text{[1]} \{rule$_1$\} \\
 \text{[2]} \{rule$_2$\} \\
... \\
Query: \{\textit{query}\}. \\
Rank the \{\textit{num}\} rules above based on their relevance to the query. All the rules should be included and listed using identifiers, in descending order of relevance. The output format should be [] > [], e.g., [2] > [1], Only respond with the ranking results, do not say any word or explain.
\\
\bottomrule
\end{tabularx}

\end{table}

\section{Experiment setting}
\label{experiment setting}

\paragraph{Test Benchmark} 
RuleBench \cite{sun2024beyond} evaluates the reasoning capabilities of large language models (LLMs) under a given set of rules. Building on the foundation of RuleBench, we consolidate all rules within the entire test set into a comprehensive rule library and used the original questions as queries. We establish both a natural language-based and a formal language-based rule library to assess the impact of different rule formats on retrieval performance.  We select two synthetic datasets, Clutrr \cite{sinha-etal-2019-clutrr} and ULogic \cite{wang-etal-2024-llms}, as well as a real-world dataset, CAIL2018 \cite{xiao2018cail2018largescalelegaldataset} from RuleBench for our evaluation.

\paragraph{Metrics}
We employ Recall@1, Recall@5, and Recall@10 to evaluate retrieval performance, and use the Match metric \cite{rau2024bergenbenchmarkinglibraryretrievalaugmented} to assess reasoning performance. Specifically, if the golden answer appears in the final answer generated by the LLM, then it is considered correct.

\paragraph{Implementation Details}
We use Pyserini \cite{lin2021pyserini} to implement the BM25 retriever and employ bge-base-en \cite{bge_embedding} as the dense encoder. For self-induction, rule relevance re-estimation, and final reasoning, we utilize the gpt-4o \cite{gpt4o} and Qwen2.5 \cite{qwen2.5} series (7B and 72B) as the tested LLMs. We leverage VLLM \cite{kwon2023efficient} to accelerate inference. For SIAR, we get top-10 rules for retrieval performance evaluation and use the top-1 rule for reasoning evaluation. For SIAR-R$^3$, we get top-20 rules by SIAR, and use R$^3$ to get the top-10 relevant rules for retrieval evaluation and use the top-1 rule for reasoning evaluation.

\section{Retrieval performance with rule library in Formal Language}
\label{formal retrieval results}
Due to the api cost, we do not test the gpt-4o performance on Formal Language rules.
We show the results in Talbe \ref{tab:formal_retrieval}.
\textbf{\begin{table*}[t]
\centering
\tabcolsep=15pt
\resizebox{1\linewidth}{!}{
\begin{tabular}{lccccccccc}
\toprule
\multirow{2}{*}{} & \multicolumn{3}{c}{\textbf{CLUTRR}} & \multicolumn{3}{c}{\textbf{ULogic}}  & \multicolumn{3}{c}{\textbf{CAIL2018}}  \\
\cmidrule(l){2-10}
 & R@1 & R@5  & R@10 & R@1 & R@5  & R@10 & R@1 & R@5  & R@10  \\
\cmidrule{1-10}
\textbf{sparse retrieval (BM25)} \\
\cmidrule(l){1-1}
\texttt{vanilla retrieval} &  6.58 & 16.60 & 24.43 & 22.41 & 44.10 & 51.57 & 22.89 & 42.77 & 53.01  \\
\cmidrule{1-10}
\texttt{Qwen2.5-72B-Instruct}  \\
+ SIAR (w/ SI) &  6.58 & 18.89 & 27.67 & 45.90 & 68.67 & 75.30 & 79.52 & 93.37 & 95.18  \\ 
+ SIAR (w/ SI + input) & 10.01 & \textbf{23.76} & \textbf{31.97} & 51.57 & 73.49 & 80.24 & 59.04 & 83.13 & 89.15 \\ 
+ SIAR-R$^3$ (w/ SI) &  10.21 & 18.80 & 26.81 & 75.54 & 80.12 & 80.36 & \textbf{83.13} & \textbf{95.78} & \textbf{95.78} \\ 
+ SIAR-R$^3$ (w/ SI + input) & \textbf{11.07} & 23.00 & 30.25 & \textbf{80.60} & \textbf{85.42} & \textbf{85.66} & 83.13 & 93.37 & 93.37 \\ 
\cmidrule{1-10}
\texttt{Qwen2.5-7B-Instruct}  \\
+ SIAR (w/ SI) &  2.39 & 6.34 & 10.21 & 47.47 & 68.67 & 76.27 & 69.88 & 88.55 & 93.37  \\ 
+ SIAR (w/ SI + input) & \textbf{7.16} & \textbf{16.89} & \textbf{23.57} & 50.48 & 73.49 & 81.08 & 46.99 & 72.89 & 82.53 \\ 
+ SIAR-R$^3$ (w/ SI) &  1.34 & 5.25 & 9.73 & 72.17 & 81.33 & 82.17 & \textbf{78.31} & \textbf{92.17} & \textbf{93.37} \\ 
+ SIAR-R$^3$ (w/ SI + input) & 2.67 & 11.07 & 17.84 & \textbf{75.90} & \textbf{85.66} & \textbf{86.02} & 72.29 & 87.35 & 89.76 \\ 
\cmidrule{1-10}
\textbf{dense retrieval (bge)} \\
\cmidrule(l){1-1}
\texttt{vanilla retrieval} &  2.86 & 8.59 & 12.79 & 18.31 & 43.98 & 55.66 & 1.81 & 7.83 & 14.46  \\
\cmidrule{1-10}
\texttt{Qwen2.5-72B-Instruct}  \\
+ SIAR (w/ SI) &  8.30 & \textbf{20.90} & \textbf{27.10} & 76.50 & 90.72 & 94.46 & 40.96 & 60.24 & 64.46  \\ 
+ SIAR (w/ SI + input) & 4.29 & 10.59 & 17.37 & 30.0 & 60.24 & 69.88 & 6.02 & 19.28 & 33.13 \\ 
+ SIAR-R$^3$ (w/ SI) & \textbf{8.87} & 18.70 & 25.48 & \textbf{89.75} & \textbf{95.90} & \textbf{96.02} & \textbf{62.65} & \textbf{71.69} & \textbf{71.69} \\ 
+ SIAR-R$^3$ (w/ SI + input) & 6.97 & 14.98 & 19.47 & 73.25 & 80.36 & 80.60 & 37.35 & 42.17 & 42.77 \\ 
\cmidrule{1-10}
\texttt{Qwen2.5-7B-Instruct}  \\
+ SIAR (w/ SI) &  2.96 & 8.59 & 11.93 & 78.43 & 92.17 & 95.54 & 24.70 & 44.58 & 54.22  \\ 
+ SIAR (w/ SI + input) & \textbf{3.81} & \textbf{9.64} & \textbf{15.08} & 31.08 & 60.36 & 69.76 & 8.43 & 19.88 & 29.52 \\ 
+ SIAR-R$^3$ (w/ SI) & 1.43 & 6.58 & 10.30 & \textbf{88.31} & \textbf{96.98} & \textbf{97.23} & \textbf{55.42} & \textbf{64.46} & \textbf{64.46} \\ 
+ SIAR-R$^3$ (w/ SI + input) & 3.05 & 7.35 & 11.93 & 69.40 & 80.24 & 80.48 & 33.73 & 40.96 & 41.57 \\ 
\bottomrule
\end{tabular}
}
\caption{Performance of different methods with rule library in \textit{Formal Language}. We use Recall@1, Recall@5 and Recall@10 as the retrieval metrics.}
\label{tab:formal_retrieval}
\end{table*}}

\section{More analysis on retrieval results.}
\label{retrieval analysis}

\paragraph{w/ SI is suitable for dense retrieval, while w/ SI + input is suitable for sparse retrieval.}
Under the same conditions, we observe that when using SI as the query, dense retrieval typically outperforms sparse retrieval. Conversely, when using SI+input as the query, sparse retrieval tends to perform better than dense retrieval. This difference can be attributed to the nature of the two retrieval methods. Dense retrievers map both the query and the rule into a unified vector space to measure semantic similarity, whereas sparse retrievers rely on keyword matching. When SI+input is used as the query, it can disrupt the semantic coherence of the SI, while the rules in the library remain intact, resulting in a decrease in similarity within the vector space. As a result, dense retrieval is more effective when SI alone is used as the query. In contrast, for sparse retrieval, if the query contains keywords from the target rule, it can augment the SI, thus increasing the BM25 score between the SI and the rule. This makes sparse retrieval more suitable when SI+input is used as the query.

\paragraph{Rule library suits more in the format of \textit{Natural Language}.}
By comparing Table \ref{tab:natural_retrieval} and Table \ref{tab:formal_retrieval}, we observe that SIAR and SIAR-R$^3$ perform better when retrieving from the natural language rule base. Rules expressed in formal language are more abstract and harder to interpret, making it more challenging for the LLM to perform self-induction and assess relevance. Poor self-induction and relevance reestimation by the LLM can therefore degrade retrieval quality.

\section{Ablation results on types of retrievers and the number of rules.}
We show the ablation results in Table \ref{tab:retrievers} and Table \ref{tab:nums}.
\label{abl: app}
\begin{table*}[h]
    \centering
    \resizebox{1\linewidth}{!}{
    \begin{tabular}{lccccccccc}
    \toprule
    & \multicolumn{3}{c}{\texttt{Sparse Retriever}} & \multicolumn{3}{c}{\texttt{Dense Retriever}} & \multicolumn{3}{c}{\texttt{LLM retriever}} \\
    \cmidrule{2-10}
    & R@1 & R@5 & R@10 & R@1 & R@5 & R@10 & R@1 & R@5 & R@10 \\
    \midrule
    \textit{w/} \texttt{Qwen2.5-7B-Instruct} \\
    vanilla retrieval & 68.91 & 85.42 & 92.29 & 30.36 & 58.43 & 71.45 & 74.94 & 95.18 & 97.83 \\
    vanilla retrieval + R$^{3}$ & 86.99 & 93.98 & 94.70 & 70.12 & 79.88 & 80.48 & 86.99 & 97.83& 98.67 \\
    SIAR (w/ SI) & 60.48 & 83.86 & 90.00 & 71.57 & 90.96& 95.54 &78.07& 95.06& 97.47\\
    SIAR (w/ SI + input) & 76.14& 90.24 & 95.18 & 42.89 & 71.93 & 83.01 & 86.39 & 98.67 & 99.88 \\
    SIAR-R$^3$ (w/ SI) & 84.81 & 93.13 & 93.37 & \textbf{87.59}& \textbf{96.39}& \textbf{97.11}& 88.91& 98.19& 98.67 \\
    SIAR-R$^3$ (w/ SI + input) & \textbf{88.67}& \textbf{96.14}& \textbf{96.50}&76.75& 87.95& 88.92& \textbf{90.84}& \textbf{98.91}& \textbf{99.28} \\
    \bottomrule
    \end{tabular}
    }
    \caption{Retrieval performance with different types of retrievers.}
    \label{tab:retrievers}
\end{table*}

\begin{table}[h]
    \centering
    \resizebox{1\linewidth}{!}{
    \begin{tabular}{lcc}
    \toprule
     & \texttt{Original}  &	\texttt{Original + Counterfactual}  \\
    \midrule
    \textit{w/ sparse retrieval} \\
    vanilla retrieval & 68.91/ 85.42/ 92.29& 49.04/ 79.40/ 85.30 \\
    vanilla retrieval + R$^3$ & 86.99/ 93.98/ 94.70& 79.64/ 90.48/ 91.69  \\
    SIAR (w/ SI) & 60.48/ 83.86/ 90.00& 59.16/ 77.11/ 84.34 \\
    SIAR (w/ SI + input) & 76.14/ 90.24/ 95.18 & 72.53/ 87.71/ 91.08  \\
    SIAR-R$^3$ (w/ SI) & 84.81/ 93.13/ 93.37 &  81.20/ 90.00/ 90.84 \\
    SIAR-R$^3$ (w/ SI + input) & \textbf{88.67} / \textbf{96.14} / \textbf{96.50} & \textbf{86.99} / \textbf{94.10} / \textbf{94.94}  \\
    \midrule
    \textit{w/ sparse retrieval} \\
    vanilla retrieval & 30.36/ 58.43/ 71.45 & 20.48/ 49.52/ 60.60 \\
    vanilla retrieval + R$^3$ & 70.12/ 79.88/ 80.48& 56.02/ 69.64/ 70.84  \\
    SIAR (w/ SI) & 71.57/ 90.96/ 95.54& 67.83/ 87.47/ 92.53 \\
    SIAR (w/ SI + input) & 42.89/ 71.93/ 83.01& 32.53/ 62.77/ 73..49  \\
    SIAR-R$^3$ (w/ SI) & \textbf{87.59} / \textbf{96.39} / \textbf{97.11} & \textbf{84.33} / \textbf{94.57} / \textbf{95.42} \\
    SIAR-R$^3$ (w/ SI + input) & 76.75/ 87.95/ 88.92 & 66.87/ 80.48/ 82.65  \\
    \bottomrule
    \end{tabular}
    }
    \caption{Retrieval performance with different number of rules. Original represents the rule set used in the previous section. Counterfactual represents the addtional rules we select from the RuleBench.}
    \label{tab:nums}
\end{table}

\end{document}